\documentclass{article}
\usepackage{templates/arxiv}

\usepackage[english]{babel}
\usepackage{lipsum}

\usepackage{amsmath}
\usepackage{amsfonts}
\usepackage{graphicx}
\usepackage{subcaption}
\usepackage{makecell}
\usepackage{multirow}
\usepackage{comment}
\usepackage{indentfirst}
\usepackage[ruled]{algorithm2e}
\usepackage[svgnames,xcdraw,table]{xcolor}
\usepackage[colorlinks=true, allcolors=blue]{hyperref}

\addto\extrasenglish{  
  
}

\usepackage{setspace}

\newcommand{\abs}[1]{\left|#1\right|}

\newcommand{\pardelim}[1]{\left(#1\right)}
\newcommand{\sqrdelim}[1]{\left[#1\right]}
\newcommand{\angdelim}[1]{\left\langle#1\right\rangle}
\newcommand{\crldelim}[1]{\left\{#1\right\}}
\newcommand{\rev}[1]{}
\newcommand{\cmt}[1]{#1}
\newcommand{\ie}{\textit{i.e.}~}
\newcommand{\eg}{\textit{e.g.}~}

\title{Shape irregularity of Life-Like Network Automaton rules as an indicator of classification performance}
\renewcommand{\shorttitle}{Shape irregularity of LLNA rules as an indicator of classification performance}

\author{
    \href{https://orcid.org/0000-0002-7881-8414}{\includegraphics[scale=0.06]{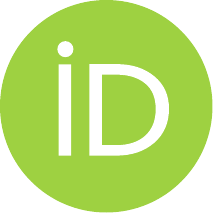}\hspace{1mm}Lucas C.~S. Oliveira}\\
	Institute of Mathematics and Computer Science\\
    University of São Paulo\\
	São Carlos, 13566-590, Brazil\\
	\texttt{l.caldeira@usp.br}\\
	\And
    \href{https://orcid.org/0000-0001-8467-734X}{\includegraphics[scale=0.06]{templates/orcid.pdf}\hspace{1mm}Michiel Rollier}\\
	BionamiX, Dept. of Data Analysis and Mathematical Modelling\\
    Ghent University\\
	Ghent, 9000, Belgium\\
	\texttt{michiel.rollier@ugent.be}\\
	\And
    \href{https://orcid.org/0000-0003-4084-9992}{\includegraphics[scale=0.06]{templates/orcid.pdf}\hspace{1mm}Jan Baetens}\\
	BionamiX, Dept. of Data Analysis and Mathematical Modelling\\
    Ghent University\\
	Ghent, 9000, Belgium\\
	\texttt{jan.baetens@ugent.be}\\
	\And
    \href{https://orcid.org/0000-0002-2945-1556}{\includegraphics[scale=0.06]{templates/orcid.pdf}\hspace{1mm}Odemir M.~Bruno}\\
	AICS, São Carlos Institute of Physics\\
    University of São Paulo\\
	São Carlos, 13566-590, Brazil\\
	\texttt{bruno@ifsc.usp.br}\\
}
\date{October 5, 2026}

\hypersetup{
    pdftitle={Shape irregularity of LLNA rules as an indicator of classification performance},
    pdfsubject={cs.AI},
    pdfauthor={Lucas C.~S. Oliveira, Michiel Rollier, Jan Baetens, Odemir M.~Bruno},
    pdfkeywords={network automata, network classification, cellular automata, the Game of Life, metaheuristics},
}

\begin{document}
\maketitle

\begin{abstract}
Complex Network (CN) classification requires high-level structural characterizations that are both scale-invariant and computationally efficient. Methods based on Life-Like Network Automata (LLNA) offer an interesting way to extract network descriptors by leveraging emergent temporal patterns without requiring provided features, but their efficacy is bottlenecked by a high-cost combinatorial optimization problem: the selection of the automaton transition rule. While current literature relies on exhaustive searches that are unfeasible for large-scale applications, this work reveals that the rule space is fundamentally structured by a property we term ``jaggedness'', that quantifies the resemblance of a LLNA transition function with a sawtooth shape. We demonstrate that this metric acts as a theoretical proxy for chaoticity and sensitivity -- properties essential for generating discriminative dynamic behaviors among network categories. Moreover, we introduce a heuristic search strategy that uses jaggedness to guide the rule selection. Experimental results show that our approach achieves classification accuracies within 5\% of the global optimum while reducing computational overhead by 90\% compared to exhaustive approach. Our findings provide a novel, efficient, framework for optimizing automata-based methods for pattern recognition.
\end{abstract}

\keywords{network automata \and network classification \and cellular automata \and the Game of Life \and metaheuristics}

\section{Introduction}

Network science is an interdisciplinary field that mainly encompasses graph theory and statistical physics, allowing the study of complex and nonlinear phenomena in their entirety \cite{Barabasi2016:NetworkScienceBook}. Due to its great power of generalization, modeling phenomena as a graph or a complex network (CN) has been used in diverse domains -- social networks \cite{Sheridan2003:SocialNetworks}, metabolic chains \cite{Jeong2000:MetabolicNetworksOrganization}, epidemics \cite{Scabini2021:EpidemicsInterationLayers}, etc. -- and in diverse applications -- network classification \cite{Errica2019:FairComparisonClassificationGNNs}, spatiotemporal forecasting \cite{Oliveira2024:BrazilCovidTransport}, link prediction \cite{LuZhou2011:LinkPredSurvey}, identification of vulnerabilities and anomalies \cite{Liu2022:AnomalyDetection}, and others \cite{Costa2011:CNApplicationsSurvey}. 

In order to study CNs and capture more meaningful information that underlies their topology, a common approach is to compute some set of metrics that summarize certain structural aspects at node, edge and/or graph-level \cite{Costa2007:CNMeasuresSurvey}. Yet, in cases as the network classification problem, where the objective is to identify the class of a given CN \cmt{(\eg finding the taxonomic group of a metabolic network or if a given chemical compound is carcinogenic or not),} they may not be sufficiently informative or be incomplete/partially unavailable \cite{Miranda2016:ExploringTEPsLLNA}.

A method that has been successfully applied in this context is the Life-Like Network Automaton (LLNA) \cite{Miranda2016:ExploringTEPsLLNA, Machicao2018:AutorshipLLNA, Ribas2020:BinaryPatternsLLNA, Zielinski2022:ClassificationDTEPsLLNA}. Taking inspiration from Conway's Game of Life \cite{Gardner1970:SolitaireGameLife}, it relies on network topology as source of dynamical information, differing from other approaches that use only structural information \cite{Shervashidze2011:WLGraphKernels, Yanardag2015:DeepGraphKernels} and/or depend on detailed feature vectors provided beforehand \cite{Zhang2018:End2EndGraphClassification, Ying2018:GraphDiffPooling}. In summary, the LLNA generates an initial configuration for the nodes and evolves them over time according to a transition function (\ie the automaton rule), whose resulting state-time diagram is used to extract rich information about network dynamics for the sake of classification.


While this method does not require any prior information besides the network topology and is both scale-invariant and computationally efficient to run, finding a good discriminant rule for the automaton is not trivial. The classification can only be performed if the chosen rule is capable of generating patterns that share common properties between the the same class (high intraclass similarity) while being sufficiently different from other classes (low interclass similarity). Yet only a small portion of the rule space present this capability. In practice, the selection of a satisfactory transition rule can be framed as a combinatorial optimization problem, which has been approached by doing an exhaustive search in the rule space \cite{Miranda2016:ExploringTEPsLLNA, Ribas2020:BinaryPatternsLLNA, Zielinski2022:ClassificationDTEPsLLNA}. While increasing the size of the rule family tends to improve classification accuracy \cite{Zielinski2024:NeighVariancyLLNA}, an exhaustive search quickly becomes computationally unfeasible.



\begin{figure}[!ht]
    \centering
    \includegraphics[width=1\linewidth]{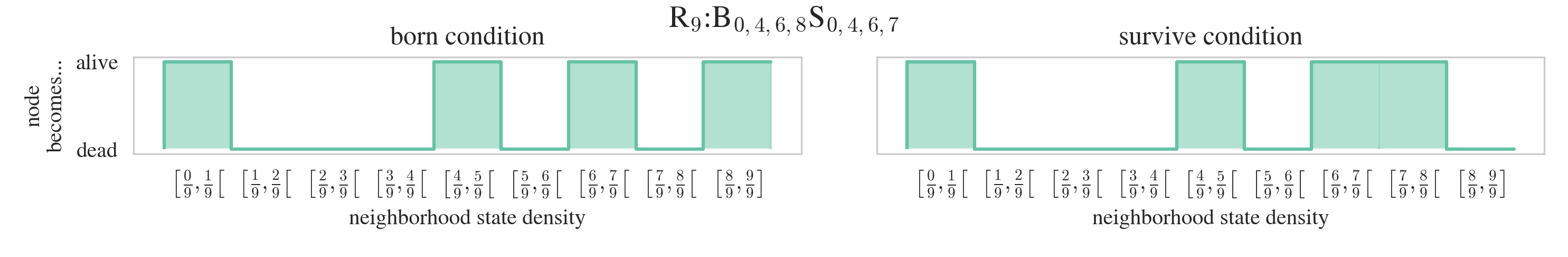}
    \\\vspace{0.5em}
    \includegraphics[width=1\linewidth]{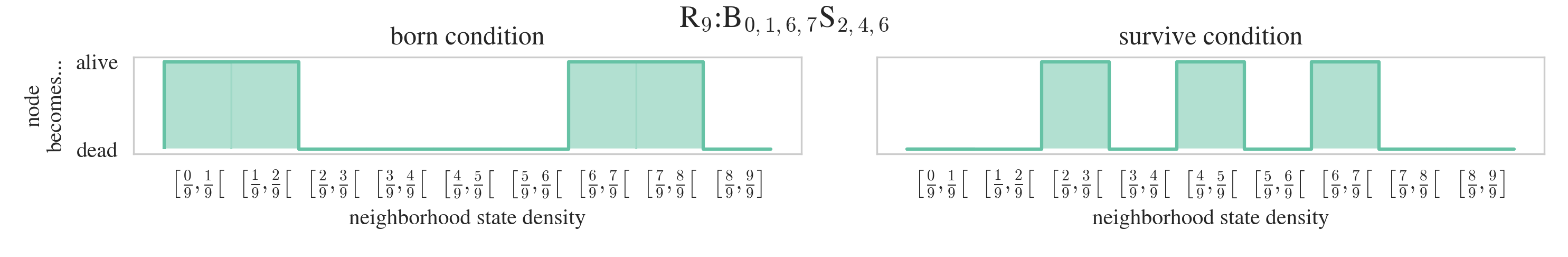}
    \caption{Best reported rules in \cite{Zielinski2022:ClassificationDTEPsLLNA} for plant metabolic chains (top) and social circles (bottom) datasets. The density intervals that can lead to alive state in both born and survive conditions are not totally contiguous, giving a jagged appearance to the transition function.}
    \label{fig:best-rules-shape}
\end{figure}

As the rule optimization is a critical bottleneck of the method, this work investigates what property characterizes LLNA rules that lead to high classification accuracies, a question that has been neglected since the method was originally proposed. After inspecting the reported best rules for some datasets, we noticed that they share certain (visual) characteristics. The set of density intervals composing the best rules presents a degree of irregularity, what in a visual perspective translates to a transition function with sawtooth shape (see \autoref{fig:best-rules-shape}). We hypothesize that this jagged appearance can be a good indicator of classification performance. 

To assess these research questions, we define a formal measure of jaggedness and investigate to what extent it can be used to organize the rule space and guide the non-exhaustive search for optimum LLNA rules. In \autoref{sec:background} we present the theoretical background of the LLNA algorithm and its use for classifying networks. In \autoref{sec:method} we formally define the jaggedness, explore how it can be used to partition the rule space, and outline a heuristic search strategy that benefits from this organization. In \autoref{sec:results} we evaluate the connection between rule jaggedness, classification performance and rule space resolution, and discuss the use of jaggedness as an indicator of complex dynamical behavior.

\section{Background}
\label{sec:background}

One of the objects of study in network theory is how the topological properties  of a CN shape its emergent dynamics. From the small-world and scale-free phenomena to hierarchy, modularity, and degree heterogeneity, structural properties -- both local and global -- influence how information and diseases spread, the resilience against random and targeted attacks, the sensitivity to small perturbations, and the emergence of collective ordered states \cite{Mitchell2009:ComplexityTourBook, Latora2017:CNPrinciplesBook}. Works such as Miranda's \cite{Miranda2016:ExploringTEPsLLNA} and Machicao's \cite{Machicao2018:AutorshipLLNA} demonstrate that it is in fact possible to characterize a network based only on its dynamical behavior, therefore allowing for pattern recognition tasks such as classification.

In this work we focus on LLNA, that generalize the born/survive dynamics from Conway's Game of Life cellular automaton (CA) to a network topology. Below we describe how one can define a CA for arbitrary networks and utilize it for characterization and classification purposes.

%
%
%

\subsection{Life-like NAs}

A \textit{network automaton} (NA) is a formalism that adapts the transition logic of a CA to the non-regular topology of networks. In other words, it uses the graph as ``tessellation'', the nodes as cells and their edges as the neighborhood structure, with the transition rule acting as the criterion for cell's state update according to its current state and neighboring cells' configuration. In his influential book \cite{Watts1999:SmallWorldsDynamics}, Watts discussed the computation of a CA in Small-World networks for density classification and synchronization tasks. The works of Marr and Hütt \cite{Marr2009:OuterTotalGraphCA, Marr2012:TopologicalPropGraphCA} and Baetens and Baets \cite{Baetens2011:TopologicalSensitivity, Baetens2012:IrregTesselationCA} approach the impact of the topology in the resulting dynamics, and propose different ways to implement CA transition rules on networks.

Moreover, it is not trivial to adapt the local transition function from CAs, due to network's intrinsic lack of ordering between nodes and unbounded number of neighbors. The former is commonly solved by using totalistic rules \cite{Rollier2024:TaxonomyCA}, but there is no standard approach for the latter, with some restricting the neighborhood size to a maximum value and others implementing a degree-based normalization, which transforms the state count operation into state density computing. As an example of a NA with a density-based transition function, we have the the so called LLNA \cite{Miranda2016:ExploringTEPsLLNA}, whose state space and transition function are inspired by the Game of Life \cite{Gardner1970:SolitaireGameLife}. It works as follows.

Each node $v_i$ of the network $\mathcal{G}$ can take one of two states, \textit{alive} (1) or \textit{dead} (0). Given the initial configuration of the network $\mathbf{s}^0 = \sqrdelim{s^0_i} \in \mathcal{S}^N$, where $s^t_i$ is the state of $v_i$ at time $t$ and $N$ is the number of nodes, the state of each $i$-th node at the next time step $t+1$ is determined by its current state $s^t_i$ and the states of its neighboring $\{s^t_j \,|\, v_j \in \mathcal{N}_i\}$. The function $\varphi$ that maps $s^t_i$ to $s^{t+1}_i$ is called transition rule, and operates according to birth and survival criteria.

\begin{figure}[!ht]
    \centering
    \includegraphics[width=0.7\linewidth]{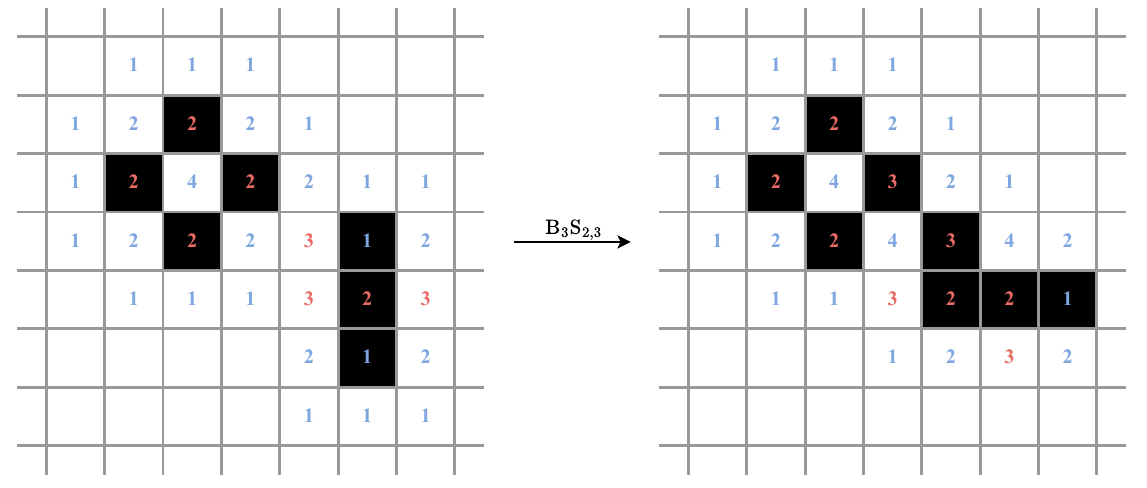}
    \\\vspace{0.5em}
    \includegraphics[width=0.7\linewidth]{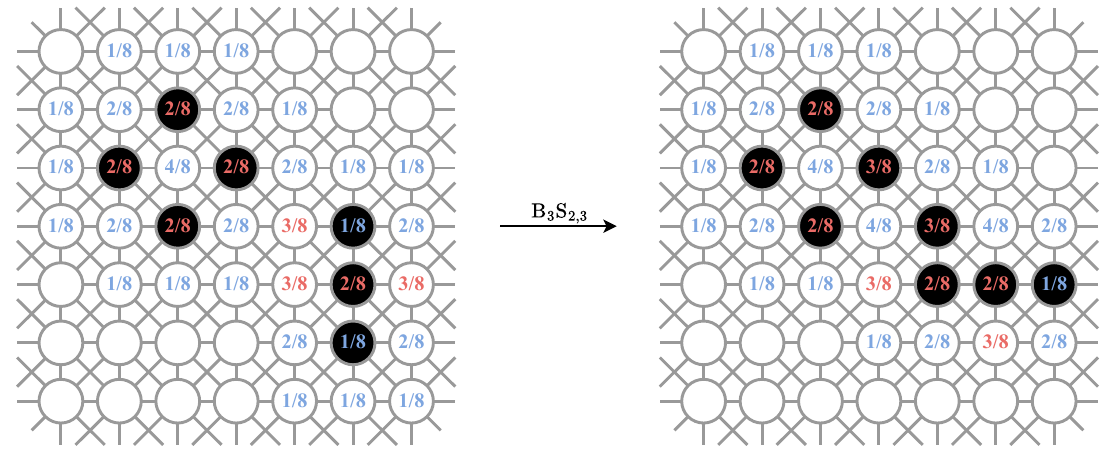}
    \\\vspace{0.5em}
    \includegraphics[width=0.7\linewidth]{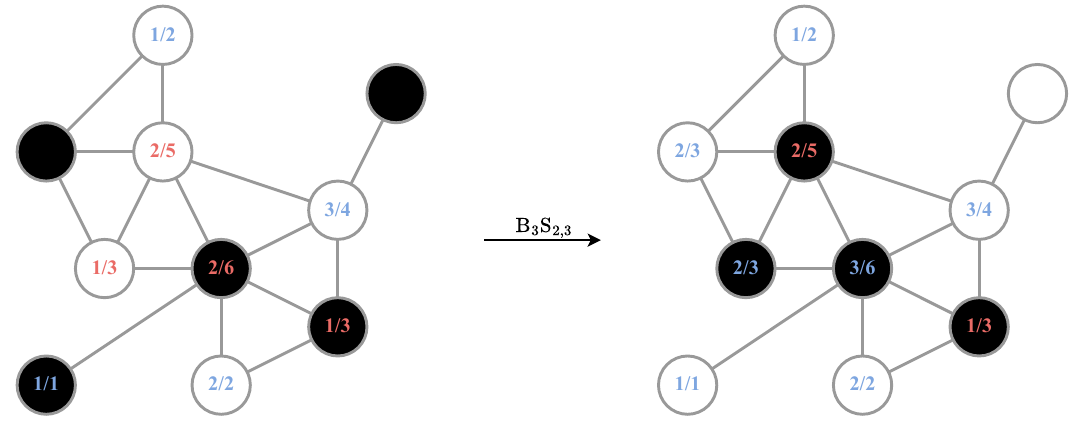}
    \caption{Examples of the Game of Life transition rule operating on an infinite regular 2-D grid with Moore's neighborhood (top) with the counting of states, on a network topology equivalent to the Moore's neighborhood (middle) using density-based operation, and on a finite arbitrary network topology (bottom). Living cells/nodes are present in black, and dead ones in white, with their amount of living neighbors denoted (zeros are omitted).}
    \label{fig:game-of-life-examples}
\end{figure}

For example, the transition rule of the original Game of Life states that any dead cell with exactly 3 living neighbors or any live cell with 2 or 3 living neighbors, out of its 8 neighbors, will have its next state set to alive, otherwise it will stay/become dead. In the case of LLNA, the equivalent rule is obtained by computing the density of living neighbors and checking if the density falls into $\left[ \frac{3}{9}, \frac{4}{9} \!\right[$, for the born criterion, or $\left[ \frac{2}{9}, \frac{3}{9} \!\right[ \,\cup \left[ \frac{3}{9}, \frac{4}{9} \!\right[$, for the survive criterion. \autoref{fig:game-of-life-examples} illustrates this logic, where the count/density of living neighbors that will trigger the birth-and-survival rule is marked in red.

Formally, the LLNA is a Boolean outer-totalistic NA defined by the quadruple $\angdelim{\mathcal{G}, \mathcal{S}, \mathbf{s}^0, \varphi^\alpha}$, with the transition rule $\varphi^\alpha = \varphi$ operating on a uniform partition with cardinality $\alpha$ of the unit interval. Let $\mathcal{H}_k^\alpha$ be the $k$-th subset of the uniform partition, 
\begin{equation}
    \mathcal{H}_k^\alpha = 
    \begin{cases}
        \left[ \frac{k}{\alpha}, \frac{k+1}{\alpha} \right[ &,\; k = 0, \dots, \alpha-2; \\
        \left[ \frac{k}{\alpha}, \frac{k+1}{\alpha} \right] &,\; k = \alpha-1. \\
    \end{cases}
\end{equation}
A \textit{born} criterion is satisfied if $s^t_i = 0$ and the fraction of living nodes in the neighborhood of $v_i$, given by $\rho^t_i = \frac{1}{d_i}\sum_{v_j \in \mathcal{N}_i} s^t_j$, falls in some interval covered by the set $B_X = \bigcup_{k \in X} \mathcal{H}_k^\alpha$, where $d_i = \abs{\mathcal{N}_i}$ is the degree of the node $i$. Analogously, a \textit{survive} criterion is satisfied when $s^t_i = 1$ and $\rho^t_i \in S_Y$, with $S_Y = \bigcup_{k \in Y} \mathcal{H}_k^\alpha$.

\begin{equation}
    s^{t+1}_i = 
    \varphi(s^t_i, \rho^t_i) =
    \begin{cases}
        1 &\text{if}\; s^t_i = 0 \;\text{and}\; \rho^t_i \in B_X; \\
        1 &\text{if}\; s^t_i = 1 \;\text{and}\; \rho^t_i \in S_Y; \\
        0 &\text{otherwise.}
    \end{cases}
    \label{eq:llna_transition_func_nodewise}
\end{equation}

Nonetheless, the parameter $\alpha$ is directly related to the term ``neighborhood size'' (of the LLNA transition rule) stated by Zielinski \cite{Zielinski2024:NeighVariancyLLNA}, as the latter is given by $\alpha - 1$. The neighborhood size of a rule refers to the number of non-zero neighborhood state configurations a given rule can distinguish. In case of the Game of Life, it stands for the Moore's neighborhood, allowing for 8 possible non-equivalent configurations of living cells and one additional with all neighbors dead. A LLNA rule that allows for $\alpha$ different density intervals therefore has a neighborhood size of $\alpha-1$. Here we prefer to name this parameter $\alpha$ as the resolution of the rule's density interval, or just rule resolution.

%
%


\subsection{Rule numbering scheme}

A LLNA rule is described by the sets of indices $X,Y \subseteq \{0, \dots, \alpha-1\}$ and the resolution $\alpha$. Bigger values for $\alpha$ allow more intricate -- and possibly better discriminant -- rules, at the cost of an exponentially increasing rule space \cite{Zielinski2024:NeighVariancyLLNA}. To avoid ambiguity when mentioning automata rules from distinct rule families $\Phi^\alpha$ (\ie the finite set of all rules with the same resolution), we propose a more complete naming convention.

The most direct way to represent a LLNA rule is by visualizing $X, Y$ as subscripts and the resolution highlighted, like $R_\alpha$:$B_XS_Y$. Another way is by encoding those indices in a more compact form. Let $\beta = \sum_{k=0}^{\alpha-1} x_{k} 2^{k}$ and $\sigma = \sum_{k=0}^{\alpha-1} y_{k} 2^{k}$ be the decimal encoding of $X$ and $Y$ respectively, with the binary coefficients $x_k = 1 \iff k \in X$ and $y_k = 1 \iff k \in Y$. Then, a rule $\varphi$ can be expressed by the triple $(\alpha, \beta, \sigma)$, or simply $\varphi^{\alpha}_{\beta, \sigma}$.

For instance, the transition rule from Game of Life is denoted by $R_9$:$B_{3}S_{2,3}$ in the explicit format and by $\varphi^{9}_{8,12}$, in the compact format.

\subsection{Network classification with LLNA}

One of LLNA’s primary applications is feature extraction from CNs (characterization task) to predict their classes (classification task), which can be as simple as distinguish between fundamental CN models or as complicated as determine the taxonomic groups of metabolic networks \cite{Miranda2016:ExploringTEPsLLNA, Machicao2018:AutorshipLLNA, Ribas2020:BinaryPatternsLLNA, Zielinski2022:ClassificationDTEPsLLNA}. The general LLNA-based classification model follows a structured pipeline composed by three modules: network automaton, feature extractor, and tabular classifier.

The \textit{automaton} generates the state-time matrix $\mathbf{S} = \sqrdelim{\mathbf{s}^0 \dots \mathbf{s}^T} \in \mathcal{S}^{N \times (T+1)}$, also called time evolution patterns (TEPs), whose shape depends on the number of time steps $T$ and the number of nodes $N$. This process is fully determined by the interaction between the network topology $\mathcal{G}$, the transition rule $\varphi$, and the initial configuration $\mathbf{s}^0$.

The \textit{extractor} maps the variable-size temporal data into a fixed-length feature vector by applying a set of feature extraction functions followed by histogram-based aggregations. 
Feature functions used in previous studies include: Shannon entropy, word entropy and Lempel-Ziv complexity \cite{Miranda2016:ExploringTEPsLLNA}; binary-pattern encoding (LLNA-BP), which performs a sliding window operation over binary sequences, analogous to a 1-D convolution \cite{Ribas2020:BinaryPatternsLLNA}; and the density/state-density TEPs, which combines the node state evolution with the corresponding density matrix \cite{Zielinski2022:ClassificationDTEPsLLNA}.

Finally, the \textit{classifier} performs the network classification, mapping the feature vector ${\Psi}$ to the network class $z$\cmt{, where $z$ can be a Boolean variable (binary classification),  -- \eg identifying if a molecule is carcinogenic -- or a non-binary categorical label (multi-class classification) -- \eg pointing which social media community generated a specific interaction networks}. The discriminant capability of the classifier is entirely dependent on the preceding modules, considering that previous works reached optimum results with shallow machine learning models, such as linear kernel SVM and $k$-NN, which do not support representation learning \cite{Hamilton2020:GRLBook}.

\begin{figure}[!ht]
    \centering
    \includegraphics[width=0.8\linewidth]{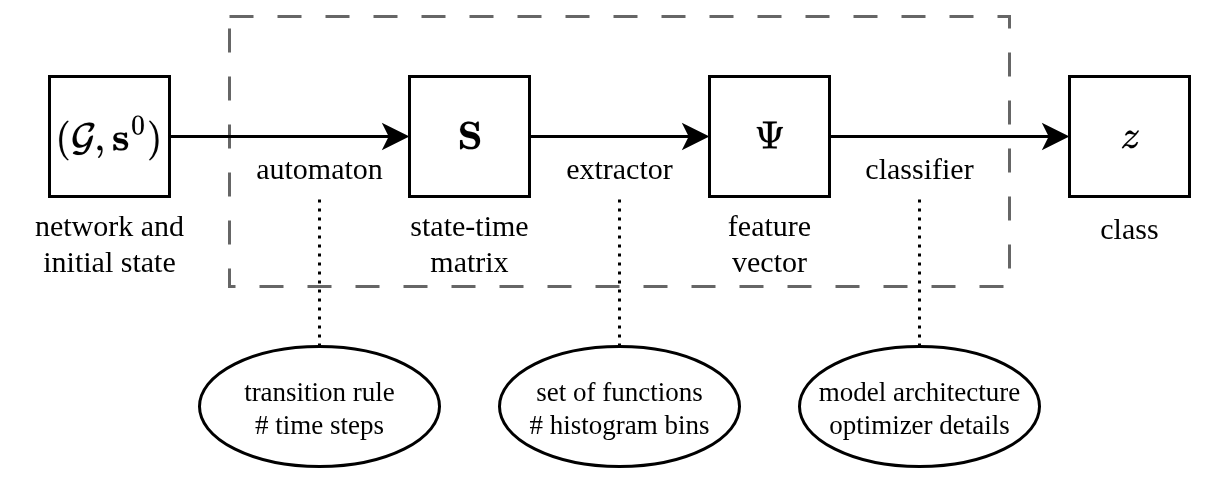}
    \caption{General network classification pipeline based on LLNA. The boxes indicate the inputs and outputs, the arrows denote the action performed by the modules, and the ellipses list their hyper-parameters. The dashed rectangle delimit the end-to-end classification model.}
    \label{fig:clf-pipeline}
\end{figure}

\autoref{fig:clf-pipeline} illustrates the classification pipeline of a dataset $\crldelim{(\mathcal{G}, z)}$. Note that the vector $\mathbf{s}^0$ does not need to be provided in advance, as it is generated at the data preprocessing step or during runtime.

A key challenge in this framework is hyper-parameter optimization. While the classifier is the only module having learnable parameters, there are several hyper-parameters in all modules to be optimized as well (see \autoref{fig:clf-pipeline}), with the automaton rule being the most critical. The tuning of the classifier must be repeated for every rule in order to find the best configuration, which has a high computational cost when considering the commonly adopted routine of a $k$-fold cross validation with a grid search for the hyper-parameters. \cmt{It becomes even worse when considering the additional task of selecting the subset of rules to explore. As the entire rule space is infinite, thus computationally intractable, previous works adopted a single rule family for the exhaustive search. But it creates what could be framed as the rule family selection problem, which consists in choosing the smallest rule family that brings the highest classification accuracy in the rule space.} Notably, subsequent studies have largely adopted the best-performing rules identified in the original work \cite{Miranda2016:ExploringTEPsLLNA} obtained at $\Phi^9$, bypassing further optimization due to its expense.

\cmt{Hence, it is crucial to identify which rules and rule families are more promising to generate good discriminant patterns, and restrict the automaton simulation and classifier training to them. The next section provide an overview of existing methods to assess the transition rules, and introduces a novel parameter followed by a navigation strategy to better explore the rule space.}

\section{Methods}
\label{sec:method}

\cmt{The possible ways a given automaton can evolve and what dynamics it can produce is strictly related to its discriminant capacity for network classification problems. For instance, a transition rule that quickly converges to an uniform global configuration may be unable to distinguish between network classes, while a rule that allow for emergent patterns and periodic structures is much more likely to be useful for this matter.}

\cmt{In this sense, Wolfram's categorization \cite{Wolfram1984:ComplexityModelCA} offers a good starting point addressing the automata dynamical behavior qualitatively. By performing a visual inspection in the generated spatiotemporal patterns of the elementary CA, it groups the transition rules according to the convergence to a state of homogeneity (class I), periodicity (class II), chaoticity (class III) or complexity (class IV). However, it is impractical to be used at large scale, as it depends on manual observation, lacks formal definition, and allows for a same rule falling in multiple categories when evolved from different initial configurations. A better approach is to use a summarizing parameter for quantitative analysis, which, ideally, should be able to partition the rule space into groups with meaningful properties \cite{Vispoel2022:GapsCA}. Below we present some well-studied parameters, their formal definition and the interpretation regarding dynamical behavior.} 

In the context of Boolean networks, the (normalized) Hamming weight accounts for the number of ones in the truth table of a Boolean function $\varphi$ with $d$ input variables, 
\begin{equation}
    \text{HW}_d(\varphi) = 
    \frac{1}{2^{d+1}} 
    \sum_{q = 0}^{d} \binom{d}{q}
    \sum_{s \in \mathcal{S}} \varphi\!\left(s, \frac{q}{d}\right),
\end{equation}
intuitively indicating the output bias of the automaton.  The outer sum and the binomial factor cycle through all neighborhood configurations and the inner part considers the activation of $\varphi$ for a given density $q/d$. \cmt{Therefore, it represents the probability of obtaining $\varphi(s, q/d) = 1$ given a random configuration in a neighborhood of size $d$.} Note that it can be easily adapted to NAs, such as $\text{HW}_\mathcal{G} = \sum_d P(d|\mathcal{G})\text{HW}_{d}$, and $\angdelim{\text{HW}}$ as the average over a set of networks $\crldelim{\mathcal{G}}$.

\cmt{This measure is equivalent to Langton's parameter $\lambda$ for two state CAs, if the rule space is restricted to strongly quiescent rules and 0 is taken as the quiescent state $s_q$. \rev{Formally, a rule $\varphi$ is said to be quiescent if a homogeneous configuration in node state $s_q$ maps to $s_q$, and it can be expanded to a strong quiescent condition by requiring that all additional homogeneous configuration also map to a fixed point, \ie $\varphi(0,0) = 0$ and $\varphi(1,1) = 1$.} Under this convention, $\lambda = \text{HW}_d(\varphi)$ gives the fraction of neighborhood configurations that do not map to the quiescent state for strongly quiescent rules, and serves as a proxy for dynamical regime: low values of $\lambda$ are associated with ordered, fixed-point dynamics, high values with chaotic behavior, and intermediate values near the phase transition with complex, class IV dynamics \cite{Langton1990:ComputationEdgeChaos}. However, $\lambda$ is known to lose discriminative power for small state spaces ($|\mathcal{S}| < 4$), which motivates the search for alternative indicators in the Boolean setting.}

Finally, the average (Boolean) sensitivity \cite{Shmulevich2004:BooleanSensitivity}, analogous to the $\mu$-sensitivity \cite{Binder1993:uSensitivityCA}, measures the expected number of output changes in response to flipping a single input variable, 
\begin{equation}
    \text{BS}_d(\varphi) = 
    \frac{1}{2^{d+1}} 
    \sum_{q = 0}^{d} \binom{d}{q}
    \sum_{s \in \mathcal{S}} \pardelim{
        \frac{\partial\varphi}{\partial s} + 
        q\frac{\partial\varphi}{\partial q^-} +
        (d-q)\frac{\partial\varphi}{\partial q^+}
    } = 
    \text{IS}_d(\varphi) + \text{NS}_d(\varphi),
\end{equation}
where $\frac{\partial\varphi}{\partial s} = \varphi\!\pardelim{s, \frac{q}{d} } \oplus \varphi\!\pardelim{s \oplus 1, \frac{q}{d}}$ and $\frac{\partial\varphi}{\partial q^\pm} = \varphi\!\pardelim{s, \frac{q}{d} } \oplus \varphi\!\pardelim{s, \frac{q \pm 1}{d}}$ are related to the concept of Boolean derivatives \cite{Vichniac1990:BooleanDerivCA} and denote the impact of a state change in the central node (identity) and its neighborhood, respectively \cite{Rollier2025:LifeOnGraphs}. Similar to $\text{HW}_d$, all possible configurations are evaluated, but now in terms of how likely the transition function is to be affected by perturbations. A high value of $\text{BS}_d(\varphi)$ represents high sensitivity to small changes, therefore correlating to chaoticity. \cmt{Furthermore, the same aggregation can be applied to compute network- and dataset-level descriptors.}

These measures are what could be called genotype parameters, as they try to describe the dynamical behavior by looking at static attributes of the automaton and the network. Opposed to the phenotype parameters, which actually assess the dynamical outcome, they offer a good glimpse of the system behavior without requiring the full simulation \cite{Vispoel2022:GapsCA, Rollier2025:LifeOnGraphs}.

Here we take one step back and disregard topology effects, investigating to what extent the transition rule itself correlates to the dynamic regime of a NA. By removing the network topology from the analysis, we can focus on the global ability of the rule to generate interesting dynamics, which is a proxy to the automaton effectiveness in classification problems. Below we introduce a novel genotype parameter called jaggedness, which considers only the shape of the transition rule.


\subsection{Measuring the rule irregularity}

\begin{figure}[!ht]
    \centering
    \includegraphics[width=0.75\linewidth]{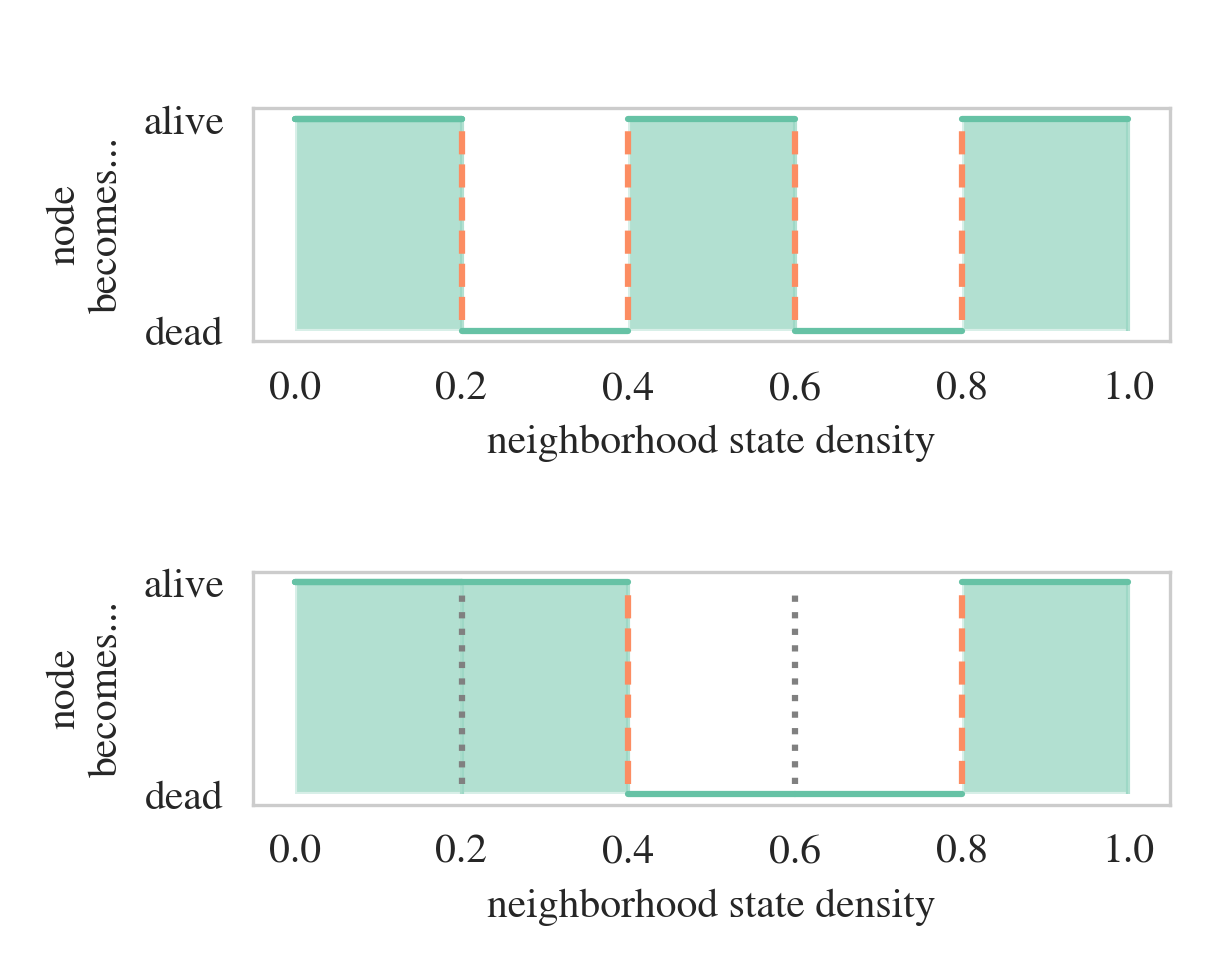}
    \caption{Diagram of the born condition for rules $\varphi^{5}_{21,4}$ (top) and $\varphi^{5}_{19,4}$ (bottom). The green areas denote the density intervals composing the rules. The vertical lines mark all possible discontinuities between subsequent intervals for $\alpha = 5$: dashed (orange) for occurring discontinuities, dotted (gray) for non-occurring ones.}
    \label{fig:jaggedness-comparison}
\end{figure}

Consider two rules $\varphi^{5}_{21,4}$ and $\varphi^{5}_{19,4}$. They are almost identical, differing only by one shifted density interval in $B_X$: while the first rule is composed by three non-adjacent intervals, $\mathcal{H}_0^5, \mathcal{H}_2^5$ and $\mathcal{H}_4^5$, the second one has two contiguous blocks, $\mathcal{H}_0^5, \mathcal{H}_1^5$ and $\mathcal{H}_4^5$ (see \autoref{fig:jaggedness-comparison}). Even though both rules cover the same number of density intervals, we can say that $\varphi^{5}_{21,4}$ is more irregular than $\varphi^{5}_{19,4}$, because its diagram is more ``jagged'' than the one of the latter.

More formally, \textit{jaggedness} can be defined as the number of discontinuities in rule diagram within $]0,1[$. Using the binary coefficients $x_k, y_k$ from the rule encoding and the XOR operator $\oplus$, we have
\begin{equation}
\begin{split}
    J(\varphi^\alpha) 
    &= \sum_{s \in \mathcal{S}} \sum_{k=0}^{\alpha-2} \left( 
        \varphi\!\left(s, \frac{k}{\alpha}\right) \oplus 
        \varphi\!\left(s, \frac{k+1}{\alpha}\right) 
    \right) \\
    &= \sum_{k=0}^{\alpha-2} (x_{k} \oplus x_{k+1}) + \sum_{k=0}^{\alpha-2} (y_{k} \oplus y_{k+1}) \;,
    \label{eq:jagged_count}
\end{split}
\end{equation}
which can be further normalized, as follows.
\begin{equation}
    \bar{J}(\varphi^\alpha) = \frac{J(\varphi^\alpha)}{2(\alpha-1)}
    \label{eq:jagged_frac}
\end{equation}

For a given resolution $\alpha$, there are $\alpha-1$ possible discontinuities for each transition criterion (born and survive) and each of them can be organized in $\binom{\alpha-1}{n}$ ways (combinations), where $J(\varphi^\alpha) = n \in \{0, \dots, 2(\alpha-1)\}$. As a combination only indicates where there is a discontinuity, without informing which adjacent interval is present in the rule, a same combination can generate two distinct rules. So, the number of rule with same jaggedness is 
\begin{equation}
    \abs{\Phi^{\alpha}_n} = 
    \sqrdelim{ 2 \binom{\alpha-1}{n} }_{\!\text{born}} 
        \times 
    \sqrdelim{ 2 \binom{\alpha-1}{n} }_{\!\text{survive}} \equiv 
    4 \binom{2(\alpha-1)}{n}
\end{equation}
and, ultimately, $\abs{\Phi^{\alpha}} = \sum_n \abs{\Phi^{\alpha}_n} = 2^{2\alpha}$.

Taking as example the top rule from \autoref{fig:jaggedness-comparison}, $\varphi = \varphi^5_{21,4} = R_5\text{:}B_{0,2,4}S_{2}$, there are 4 discontinuities in $B_X$ and 2 in $S_Y$; as a result $J(\varphi) = 6$ and $\bar{J}(\varphi) = 0.75$. Rules with this jaggedness correspond to approximately 11\% of the rule family $\Phi^5$.

In terms of the impact of changes in the node's states, the jaggedness shares the same interpretation as $\mu$-sensitivity and BS, but from a simpler and topology-independent perspective. Small values typically represent rules with boring behavior, meaning that a small changes on the neighborhood configuration rarely impacts the outcome of the transition function. On the other hand, high values may allow for chaotic behavior, quickly amplifying any perturbation and leading unpredictable trajectories.

This connection is also observed mathematically. The BS can be understood as the probability of two distinct input configurations cause different outputs multiplied by the probability of one or more nodes being simultaneously disturbed, thus transforming one input configuration into the other. Note that the latter only refers to the topology of the CN, but the former depends on the rule. Furthermore, the rule-dependent part can be separated in two effects: identity sensitivity (IS), which relates to a notion of jaggedness between transition criteria (\eg $\sum_k x_k \oplus y_k$); and neighborhood sensitivity (NS), which reflects a weighted version of the jaggedness whenever $d_i = \alpha-1$.

\begin{figure}[!ht]
    \centering
    \includegraphics[width=1\linewidth]{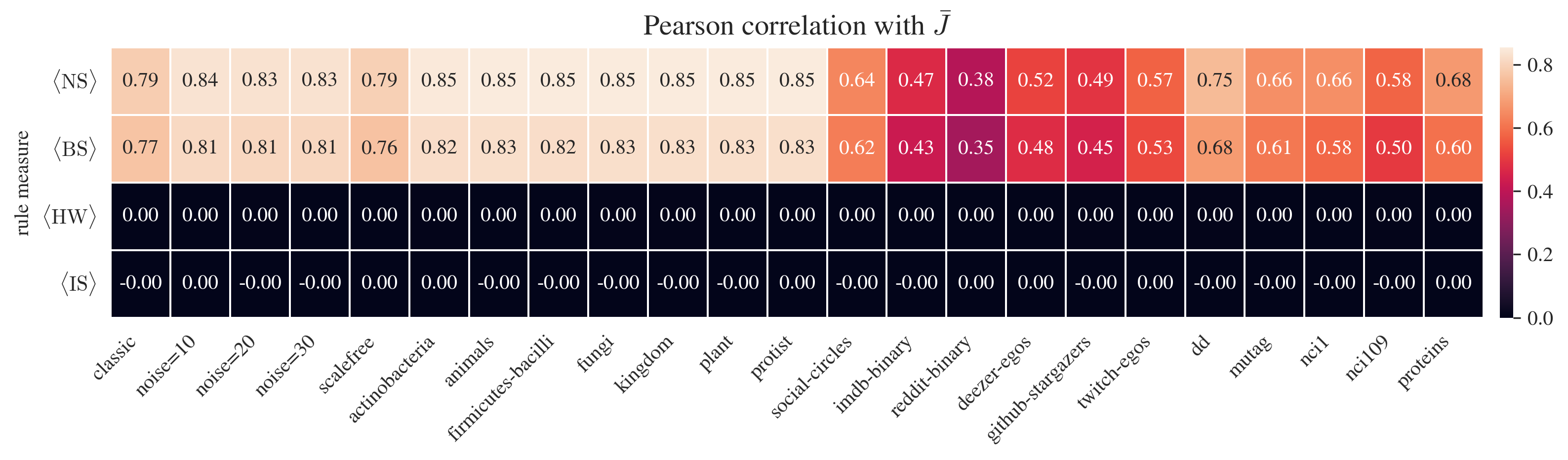}
    \caption{Correlation of the normalized jaggedness with genotype parameters, for rule families with resolution from 3 to 6.}
    \label{fig:measures_corr_jag}
\end{figure}


\subsection{Navigating through rule families}
\label{sec:navigation}

Suppose we are interested in finding the rule that maximizes the classification performance $f$. It is known that the maximum value of $f$ obtained for a given rule family, $f^{\alpha}_\text{max}$, tends to grow asymptotically as $\alpha$ increases, \cmt{with diminishing improvements after each step} \cite{Zielinski2024:NeighVariancyLLNA}. Also, in the absence of any information to guide the rule space exploration, finding this particular rule for higher resolutions is practically unfeasible. To overcome this problem, one can define a neighborhood scheme in the rule space such that two neighboring rules may have similar values of $f$, upon a (meta)heuristic search could find a local optimum $f'$ that approximates or equals the global optimum $f^*$ \cite[chapter II.4]{RusselNorvig2009:AIBook}.

\begin{figure}[!ht]
    \centering
    \includegraphics[width=0.7\linewidth]{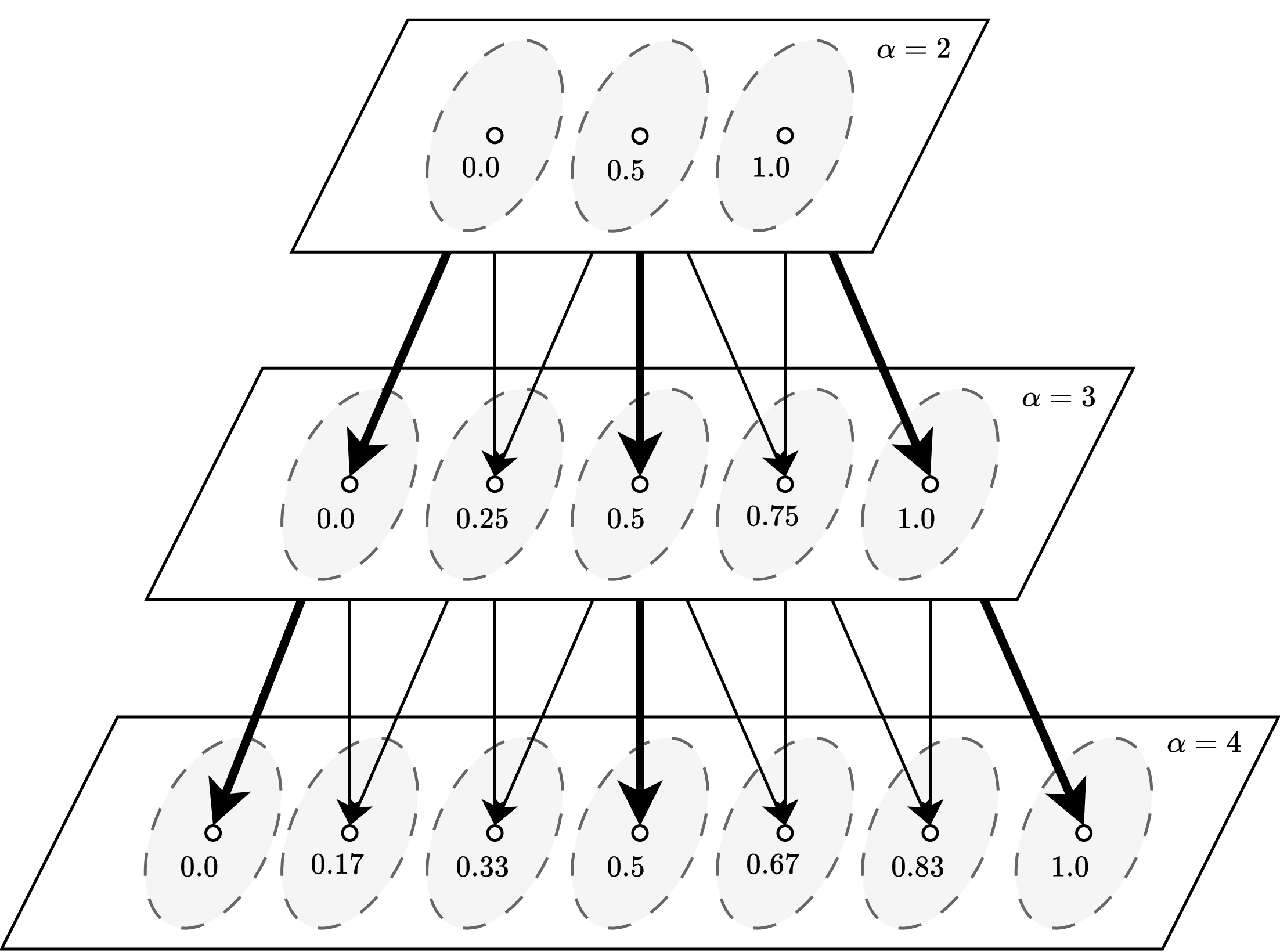}
    \caption{Navigation in LLNA rule space inspired by the ILS perturbation procedure. Each plane denotes a rule family partitioned by jaggedness. The partitions $\Phi^{\alpha}_{n}$ are represented by the ellipses with their normalized jaggedness indicated. Arrows denote the perturbation, defined as a mapping that tries to preserve $\bar{J}$. Bold arrows mark the exact mapping.}
    \label{fig:rule-space-nav}
\end{figure}

Assuming that the jaggedness is a good indicator of the rule discriminative behavior, we propose to use the rule space partitioned by $\alpha$ and $J$ as the neighborhood schema, and to apply a neighborhood navigation strategy inspired by perturbation procedure from the Iterated Local Search (ILS) metaheuristic \cite[chapter 12]{GloverKochenberger2010:MHHandbook} (see \autoref{fig:rule-space-nav}). Starting with a small family, we look for the best performing partition on average (local search step), then map it to new ones in the next family with similar normalized jaggedness (perturbation step), repeating until a stop condition is satisfied.

We now describe in what circumstances it is possible to do such optimization task with a lower computational cost than exploring an entire rule family. For that, it is important to consider two constraints. As the heuristics does not guarantee to find $f^*$, it is only worth to apply it if the total number of explored rules during the navigation between partitions is much smaller than the size of the last explored family (efficiency constraint). Additionally, the perturbation step needs to reach more than one partition in order to avoid being stuck in local optimum regions (variability constraint).

First, let us investigate how a subset of rules grows when mapped to a new subset. Let $K_\text{G}$ denote the growth rate of the size of a rule subset given a mapping scheme. In the trivial case we map the entire rule family, thus $K_\text{G}(\alpha \mapsto \alpha+1) = \abs{\Phi^{\alpha+1}}/\abs{\Phi^{\alpha}} = 4$. We can also map one partition to another with higher jaggedness in the same family, such that $K_\text{G}(n \mapsto n+1 | \alpha) = \abs{\Phi^{\alpha}_{n+1}}/\abs{\Phi^{\alpha}_{n}} = \frac{2(\alpha-1) - n}{n+1}$. Finally, we can map to a new family preserving $\bar{J}$,
\begin{equation}
    K_\text{G}((\alpha,n) \mapsto (\alpha+1, m) | \bar{J}) = 
    \frac{ \abs{\Phi^{\alpha+1}_{m}} }{ \abs{\Phi^{\alpha}_{n}} } = 
    2\alpha(2\alpha - 1)\frac{n!(2\alpha-n-2)!}{m!(2\alpha-m)!}.
    \label{eq:growing_factor_exact}
\end{equation}

There are only 3 values of $n$ that satisfy $\bar{J}(\varphi \in \Phi^{\alpha}_{n}) = \bar{J}(\varphi \in \Phi^{\alpha+1}_{m})$ and, consequently, Eq. (\ref{eq:growing_factor_exact}), being $n \in \crldelim{0, \alpha-1, 2(\alpha-1)}$; thus $\bar{J}(\varphi) \in \crldelim{0, 0.5, 1}$. When the exact mapping it is not possible, we can approximate it by identifying two consecutive integers $m^-, m^+ \in \mathbb{N}$, such that $m^+ - m^- = 1$ and $\bar{J}(\varphi \in \Phi^{\alpha+1}_{m^-}) < \bar{J}(\varphi \in \Phi^{\alpha}_{n}) < \bar{J}(\varphi \in \Phi^{\alpha+1}_{m^+})$, leading to
\begin{equation}
    K_\text{G}((\alpha, n) \mapsto (\alpha+1, \{m^-, m^+\}) | \bar{J}) = 
    \frac{ 
        \abs{\Phi^{\alpha+1}_{m^-}} + \abs{\Phi^{\alpha+1}_{m^+}} 
    }{ 
        \abs{\Phi^{\alpha}_{n}} 
    } = 
    \frac{(2\alpha + 1)2\alpha(2\alpha - 1)}{(2\alpha - n - 1)(n+1)D},
    \label{eq:growing_factor_approx}
\end{equation}
where $D = 2\alpha - n$ if $0 < n < \alpha-1$ and $D = n + 2$ if $\alpha-1 < n < 2(\alpha-1)$.

Note that the exploration variability constraint is already satisfied by the approximated mapping but not by the exact one. The expected behavior can be, therefore, achieved by including $n \mapsto m \pm 1$ in the exact operation. Equation (\ref{eq:partition-mapping}) summarizes the correct mapping between partitions,
\begin{equation}
    \Phi^{\alpha}_{n} \mapsto 
    \begin{cases}
        \Phi^{\alpha+1}_{m} \cup \Phi^{\alpha+1}_{m+1} & \text{if}\; n = 0; \\
        \Phi^{\alpha+1}_{m} \cup \Phi^{\alpha+1}_{m+1} \cup \Phi^{\alpha+1}_{m-1} & \text{if}\; n = \alpha-1; \\
        \Phi^{\alpha+1}_{m} \cup \Phi^{\alpha+1}_{m-1} & \text{if}\; n = 2(\alpha-1); \\
        \Phi^{\alpha+1}_{m^-} \cup \Phi^{\alpha+1}_{m^+} & \text{otherwise}. \\
    \end{cases}
    \label{eq:partition-mapping}
\end{equation}


Now, let $c_{\alpha+i} = \sum_{\crldelim{n}}\abs{\Phi^{\alpha+i}_{n}}$ be the cardinality of the partitions explored in $\Phi^{\alpha+i}$, subject to $\crldelim{n} \subset \crldelim{0, \dots, 2(\alpha+i-1)}$. Assuming an average $\angdelim{K_\text{G}}$ for the mapping $(\alpha+i, \{n\}) \mapsto (\alpha+i+1, \{m\})$, the exploration rate, \ie the number of rules explored after a finite sequence of mappings $\alpha \mapsto \alpha+1 \mapsto \dots \mapsto \alpha+l$ divided by the size of the final rule family, is given by the geometric series
\begin{equation}
    K_\text{E} = 
    \frac{ \sum_{i=0}^{l} c_{\alpha+i}}{ \abs{\Phi^{\alpha+l}} } =
    \frac{ \sum_{i=0}^{l} c_\alpha \angdelim{K_\text{G}}^i }{ \abs{\Phi^{\alpha+l}} } = 
    \frac{ c_\alpha }{ 4^l\abs{\Phi^{\alpha}} } \pardelim{\frac{\angdelim{K_\text{G}}^{l+1} - 1}{\angdelim{K_\text{G}} - 1}}.
    \label{eq:explor_rate}
\end{equation}

The efficiency constraint thus requires $K_\text{E} \ll 1$, which, for $\angdelim{K_\text{G}} = 4$ (worst case), is obtained whenever $c_\alpha \ll \frac{3}{4}\abs{\Phi^\alpha}$. As example, by starting the exploration at the 3 central partitions of $\Phi^4$ and considering $\angdelim{K_\text{G}} = 4$, we have $c_4 \approx 0.20 \abs{\Phi^4}$ and $\lim_{l \to \infty} K_\text{E} \approx 0.26$, which is a relevant improvement compared to the exhaustive search.

Also, it is possible to improve the theoretical computational efficiency by adopting any local search that does not fully explore the selected partitions. For instance, a simple random sampling of half of the partition rules will double the efficiency, although its impact in the overall efficacy, in terms of the resulting $f'$, is unknown. The experiments in the next section analyze some scenarios, with full and partial (sampled) exploration of the partitions, and also restricting the rules to a subset of predefined $\bar{J}$ range or only to quiescent ones.

\cmt{A final remark about the proposed optimization method is that, although being more efficient than the exhaustive search, the naive implementation of the local search as done here, consisting of a full (or partial) exploration of rule family partitions, still implies in an exponential time complexity in terms of resolution increase. For real applications, it is recommended to implement more sophisticate (meta)heuristics for the local search step, with predefined convergence and stopping criteria, thus ensuring a polynomial time complexity.}

\section{Results and Discussion}
\label{sec:results}

To evaluate the real effectiveness of the jaggedness to organize the rule space and its usefulness for guiding a search-based optimization procedure, we selected a total of 23 network datasets, half of which have been studied in LLNA-related works \cite{Miranda2016:ExploringTEPsLLNA, Ribas2020:BinaryPatternsLLNA, Zielinski2022:ClassificationDTEPsLLNA, Zielinski2024:NeighVariancyLLNA} and the other half in works applying deep learning models for graphs \cite{Yanardag2015:DeepGraphKernels, Ying2018:GraphDiffPooling, Xin2020:CNsClassificationCNN}. They vary in terms of size, connectivity, number of classes and balancing (\autoref{tab:datasets_prop}), covering a broad number of scopes\footnote{For additional explanation of the network datasets and data availability, we refer the reader to TUD and SNAP projects \cite{Morris2020:TUDataset, dataset:SNAP} and Zenodo repositories \cite{Zenodo2023:SyntheticNetworks, Zenodo2023:RealNetworks}.}:
\begin{itemize}
    \item 5 synthetic datasets generated using combinations of theoretical models (ER, BA, WS, geographic, non-linear BA), with and without noise (chance to permute an edge);
    \item 6 social networks datasets, most of them being groups of ego nets from different communities at the same social media, others being friendship bounds from distinct media;
    \item 5 bioinformatics datasets, with networks representing chemical compounds labeled according to desired properties (\textit{e.g.} carcinogenic or not);
    \item 7 metabolic networks datasets, describing interactions between molecules from taxonomic groups of living beings (\textit{e.g.} fungi, plants, fish, mammals).
\end{itemize}

\begin{table}[!ht]
    \centering
    \small
    \begin{tabular}{llccccc}
        \hline
            & \textbf{dataset}
            & \textbf{\# classes}
            & \textbf{\# nets}
            & \textbf{nets/class} 
            & \textbf{\# nodes}
            & \textbf{avg. degree} \\
        \hline
        \cite{Miranda2016:ExploringTEPsLLNA} & 
        classic                  &  4 &  11200 &           2800 &   500 -- 2000 &   4  --   16 \\
        \cite{Miranda2016:ExploringTEPsLLNA} & 
        scalefree               &  5 &    500 &            100 &          1000 &            8 \\
        \cite{Ribas2020:BinaryPatternsLLNA} & 
        noise=10\%               &  8 &    800 &            100 &          1000 &  7.6 --  8.5 \\
        \cite{Ribas2020:BinaryPatternsLLNA} & 
        noise=20\%               &  8 &    800 &            100 &          1000 &  7.6 --  8.5 \\
        \cite{Ribas2020:BinaryPatternsLLNA} & 
        noise=30\%               &  8 &    800 &            100 &          1000 &  7.6 --  8.5 \\
        \hline
        \cite{Ribas2020:BinaryPatternsLLNA} & 
        actinobacteria           &  3 &    229 &    63 --    96 &   803 -- 1533 &  3.4 -- 3.6 \\
        \cite{Ribas2020:BinaryPatternsLLNA} & 
        animals                  &  4 &     64 &             16 &  1392 -- 1919 &  3.3 -- 3.4 \\
        \cite{Ribas2020:BinaryPatternsLLNA} & 
        firmicutes-bacilli              &  4 &    454 &    86 --   143 &   530 -- 1267 &  3.2 -- 3.6 \\
        \cite{Ribas2020:BinaryPatternsLLNA} & 
        fungi                    &  4 &     68 &             17 &   624 -- 1540 &  3.0 -- 3.4 \\
        \cite{Ribas2020:BinaryPatternsLLNA} & 
        kingdom                  &  4 &    196 &             49 &   448 -- 2095 &  2.9 -- 3.5 \\
        \cite{Ribas2020:BinaryPatternsLLNA} & 
        plant                    &  3 &     33 &             11 &  1241 -- 1870 &  3.4 -- 3.5 \\
        \cite{Ribas2020:BinaryPatternsLLNA} & 
        protist                  &  4 &     28 &              7 &   598 -- 1401 &  2.9 -- 3.4 \\
        \hline
        \cite{Miranda2016:ExploringTEPsLLNA} & 
        social-circles           &  2 &   1104 &   131 --   973 &     5 -- 4938 &  0.9 --  363 \\
        \cite{Yanardag2015:DeepGraphKernels} & 
        imdb-binary              &  2 &   1000 &            500 &     5 --   38 &  1.2 --  4.6 \\
        \cite{Yanardag2015:DeepGraphKernels} & 
        reddit-binary            &  2 &   2000 &           1000 &     6 -- 1400 &  1.0 --  4.7 \\
        \cite{Rozemberczki2020:KarateClubFramework} & 
        deezer-egos              &  2 &   9629 &  4159 --  5470 &     4 --  123 &  0.9 --  7.1 \\
        \cite{Rozemberczki2020:KarateClubFramework} & 
        twitch-egos              &  2 & 127094 & 58761 -- 68333 &     5 --   33 &  1.0 --  5.7 \\
        \cite{Rozemberczki2020:KarateClubFramework} & 
        github-stargazers        &  2 &  12725 &  5917 --  6808 &     5 --  397 &  0.9 --  9.7 \\
        \hline
        \cite{Dobson2003:DistinguishingEnzymes} &
        dd                       &  2 &   1178 &   487 --   691 &    15 -- 2277 &  1.0 --  3.9 \\
        \cite{Debnath1991:MutagenicCompounds} & 
        mutag                    &  2 &    188 &    63 --   125 &     6 --   17 &  1.1 --  2.1 \\
        \cite{Wale2008:DescriptorSpacesChemicals} & 
        nci1                     &  2 &   4110 &  2053 --  2057 &     4 --   82 &  0.6 --  2.2 \\
        \cite{Wale2008:DescriptorSpacesChemicals} & 
        nci109                   &  2 &   4127 &  2048 --  2079 &     6 --   81 &  0.5 --  2.1 \\
        \cite{Borgwardt2005:ProteinGraphKernels} & 
        proteins                 &  2 &   1113 &   450 --   663 &     4 --  263 &  0.7 --  3.8 \\
        \hline
    \end{tabular}
    \caption{Properties of the selected network datasets and their respective sources, vertically grouped by scope, respectively: synthetic, metabolic, social, and bioinformatics. Ranges denote minimum and maximum values.}
    \label{tab:datasets_prop}
\end{table}

For each dataset, a small and balanced subset of a maximum of 20 networks per class was created, on which all rules from families $\alpha \in \{2, \dots, 7\}$ were applied for $T = 100$ time steps. The feature vectors were generated using density TEPs with segmentation by node degree \cite{Zielinski2022:ClassificationDTEPsLLNA}, framed in histograms with 40 bins. The overall classification pipeline was evaluated by a $5$-fold cross validation procedure repeated three times, having a Support Vector Machine with linear kernel as classifier in accordance with previous LLNA works. The chosen performance metric $f$ was the classification accuracy, \ie the fraction of network samples classified correctly. 

Below we present how the size of the rule family and the jaggedness is correlated with the classification accuracy. We applied the classification pipeline to the 23 datasets and for all of them it was possible to find at least one rule with $f \ge 60\%$, which is the minimum accuracy threshold adopted.

\subsection{Impact of resolution \rev{and quiescence} on the classification accuracy}

\begin{figure}[!ht]
    \centering
    \includegraphics[width=1\linewidth]{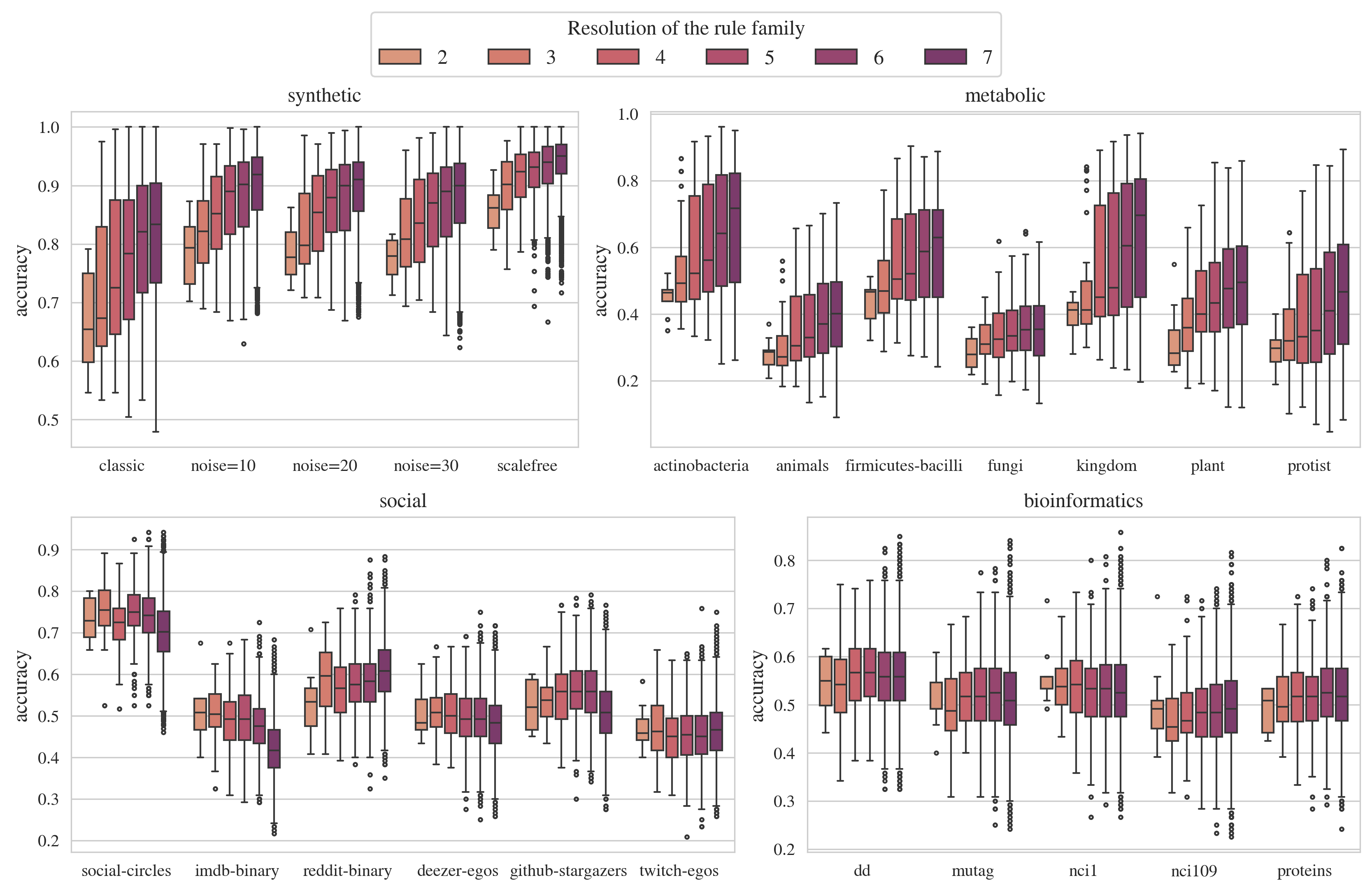}
    \caption{Box plot of the classification accuracy by rule resolution (exhaustive evaluation) for different datasets, organized in 4 groups. The box spans the inter-quartile range, with the median shown as a horizontal line and outliers as individual points.}
    \label{fig:acc_distrib_box_res}
\end{figure}

First let us analyze the accuracy for LLNA rule families $\alpha = 2, \dots, 7$ (\autoref{fig:acc_distrib_box_res}). This analysis complements a recent study \cite{Zielinski2024:NeighVariancyLLNA} by considering entire rule families instead of just the best-performing rules and by its broader scope.

In general, the maximum accuracy grows as the resolution increases asymptotically, confirming previous findings \cite{Zielinski2024:NeighVariancyLLNA}. The median accuracy, however, presents varying patterns: an increase for all the top dataset, and a constant or slightly oscillating curve for the bottom ones. Also, while some datasets reach 100\% accuracy already for small resolutions, such as any synthetic, others may demand an exploration in much larger rule families (\eg fungi networks).

\rev{Another rule property with interesting implications related to classification accuracy is the rule quiescence. Some theoretical works \cite{Langton1990:ComputationEdgeChaos, Vispoel2022:GapsCA} focus majorly or totally on quiescent rules as a way to reduce the number of possibilities to analyze and also to prioritize transition rules with more physical meaning, in a way that some conclusions about dynamical behavior only hold for this subset of transition rules. In the matter of network classification, there is no real motivation to prefer those rules besides discussing physical parallels, yet the question of its practical relevance is still valid. As seen in \autoref{fig:acc_distrib_box_qui}, while all synthetic and a few metabolic datasets (the easiest ones) show some improvement on general and minimum performances for quiescent rules others are totally unaffected by this parameter, and there is no clear pattern about the maximum classification performance.}

These results highlight the difficult trade-off between the computation cost of the rule optimization and the classification performance gain. It is not clear if a bigger rule family or a particular rule restriction would improve the current optimum accuracy, and not even if this improvement could be obtained by randomly testing new rules or just through the exhaustive exploration -- which reinforces the necessity for better, informed search strategies for the rule space. \rev{Furthermore, the differing effects of rule quiescence between synthetic and non-synthetic datasets -- and, even more critically, between trivial and non-trivial datasets -- create uncertainty about the extent to which the theoretical results apply to real-world scenarios and highlight open questions regarding the boundaries of physical implementability.}

\subsection{Impact of topology on the classification accuracy}

\begin{figure}[!ht]
    \centering
    \includegraphics[width=1\linewidth]{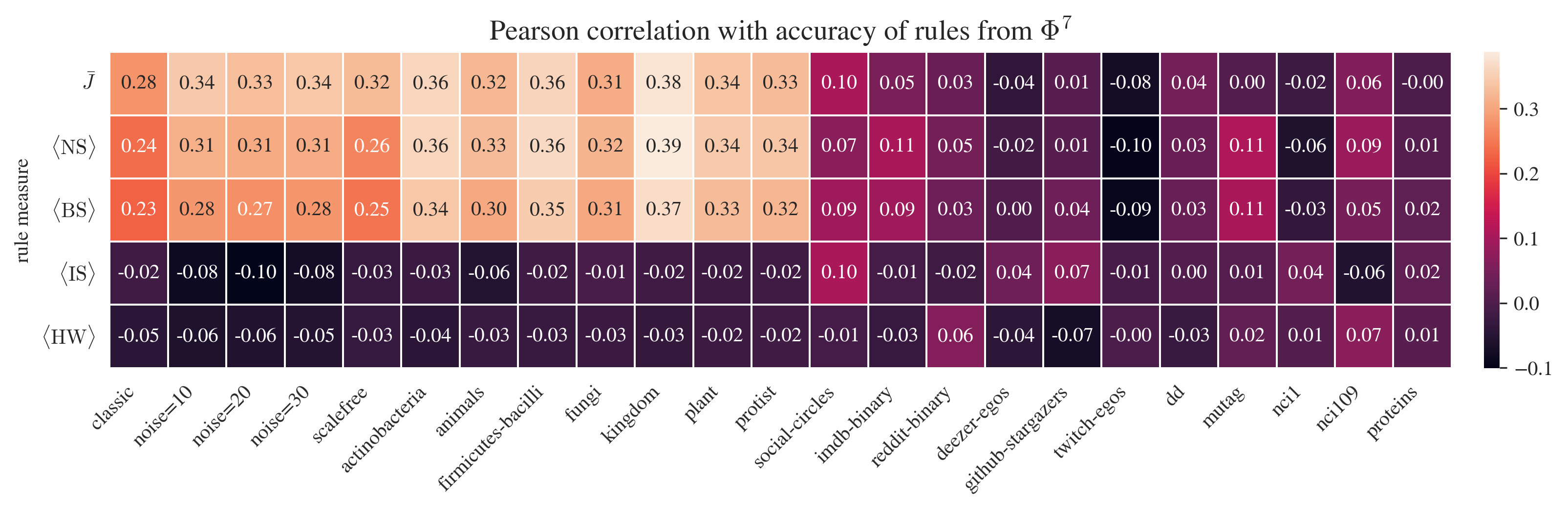}
    \caption{Correlation of the classification accuracy with the normalized jaggedness and the average Hamming weight and Boolean sensitivity, for rule family with resolution 6.}
    \label{fig:measures_corr_acc}
\end{figure}

Besides the similarities in the calculation and interpretation of jaggedness and the sensitivity measures (specially BS and its neighborhood-dependent part), it is expected that richer measures such as BS and HW perform better in capturing the dynamical behavior of the automaton, as the jaggedness is, by definition, simpler. Particularly, it cannot identify which density interval has the biggest impact on sensitivity, as it depends on the node degree distribution. It is surprising that, when comparing the correlation of the accuracy with the jaggedness and other genotype parameters over the networks of each dataset, the former exhibits a stronger correlation (\autoref{fig:measures_corr_acc}).

A possible explanation is that, in contrast with $\angdelim{\text{BS}}$, by giving equal importance to every density interval the jaggedness can be more affected by the discontinuities between intervals close to the extremities, which may be crucial for the dynamics used for classification. While for the $\angdelim{\text{HW}}$, the simple chance for a given output seems insufficient to describe the dynamical behavior related to the rules good for network classification. 

In total, 13 out of the 23 considered datasets show some degree of correlation between accuracy and normalized jaggedness, being consistent across different rule families.

\subsection{Impact of jaggedness on the classification accuracy}

\begin{figure}[!ht]
    \centering
    \includegraphics[width=1\linewidth]{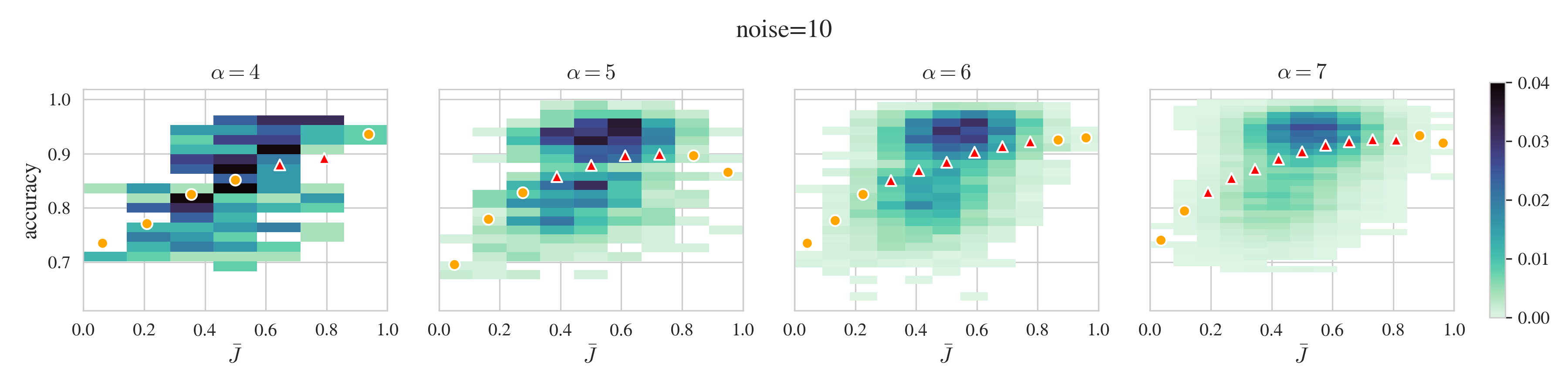}
    \includegraphics[width=1\linewidth]{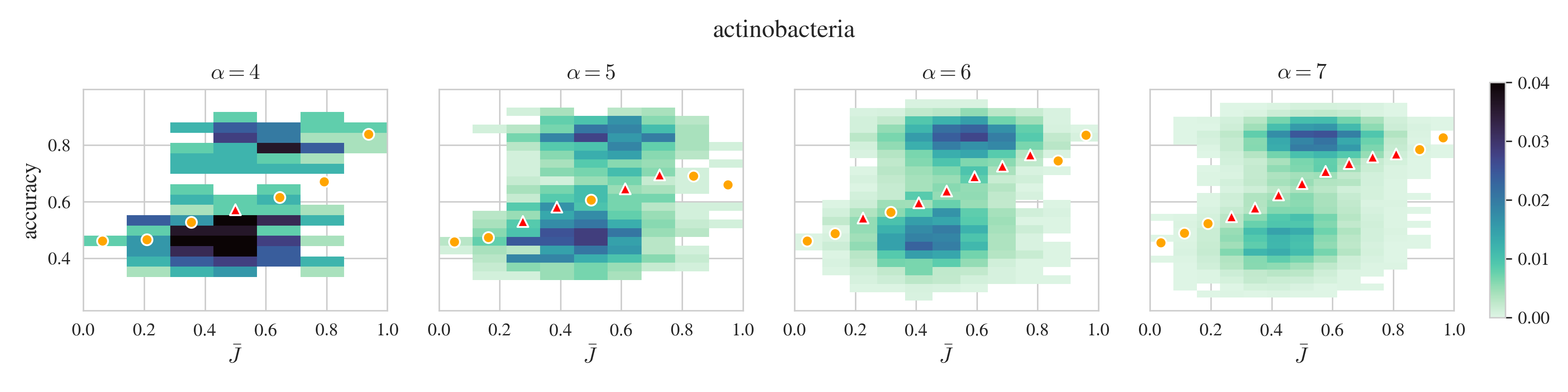}
    \includegraphics[width=1\linewidth]{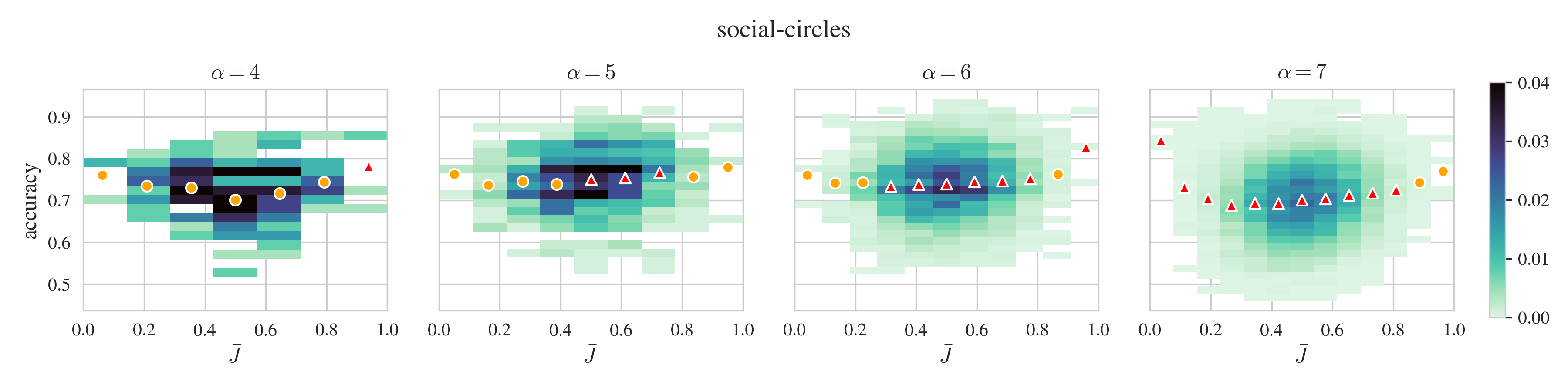}
    \includegraphics[width=1\linewidth]{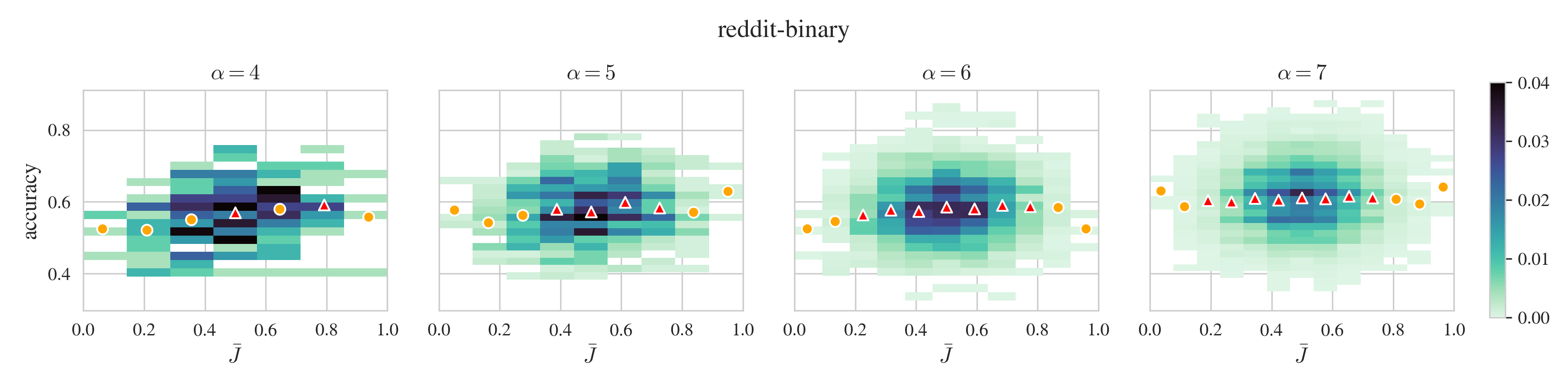}
    \caption{Accuracy distribution for the network datasets. Circles in the heatmap represent the average value for each subset of rules with same jaggedness, triangles mark the average for the subsets where there is at least one rule in the top 1\% best ones.}
    \label{fig:acc_distrib_hist}
\end{figure}

Now let us consider how the jaggedness organizes the rule families in terms of accuracy (see \autoref{fig:acc_distrib_hist}). We will limit this analysis to four datasets and resolutions 4 to 7, as the findings extend to the other datasets and larger rule families.

Starting with the synthetic noisy dataset and $\alpha = 4$, the best accuracy found is 97\% for a rule with $\bar{J} = 0.5$, but the average accuracy in this region is around 85\%, while for $\bar{J} = 1.0$ we have rules that are much better on average, and the best rule is almost as good as the global best. Proceeding to the diagrams with higher $\alpha$, the general shape remains the same: high performance variation for $\bar{J} < 0.4$, low variation for $\bar{J} \ge 0.8$ and better chances to find rules among the top 1\% when $0.5 \le \bar{J} < 1.0$.

The same analysis applies to the other three datasets. The 2-D histograms maintain their general shape over increasing resolution, and the average accuracy curve -- denoted by the circles and triangles -- quickly stabilizes. The only feature that is not present for all datasets is the skewed distribution. In the top datasets (synthetic and metabolic), there are particular values of $\bar{J}$ for which the rules accuracies vary much less and still can reach a very high classification performance. The datasets at the bottom do not have the same characteristic, being either insensitive to changes in the jaggedness or having an inconsistent sensitivity.

It is reasonable that, for those datasets where jaggedness provide some structure, the best rules appear in general at high $\bar{J}$, which are more sensitive to state density changes in the neighborhood, but rarely at $\bar{J} \approx 1.0$, as it could be too sensitive and therefore chaotic. For the other datasets, the absence of a clear pattern could either indicate that the jaggedness does not correlate with the properties that a good discriminative rule must have, or that any existing structure is only visible at bigger rule families. Particularly, the social-circles dataset was the only case with inconsistent pattern of accuracy distribution among consecutive families.

\subsection{Effectiveness of heuristic optimization}

\begin{figure}[!ht]
    \centering
    \includegraphics[width=1\linewidth]{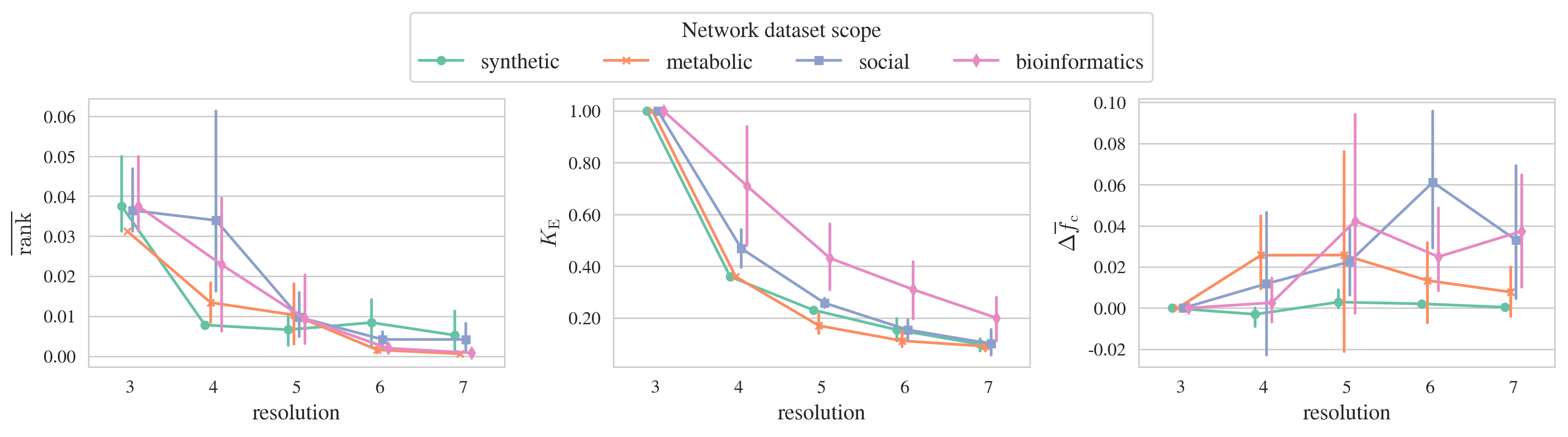}
    \caption{Performance of the heuristic search executing a full exploration of the rule family partitions, segmented by dataset scope, with 95\% confidence interval.}
    \label{fig:hs_performance_full}
\end{figure}

\begin{figure}[!ht]
    \centering
    \includegraphics[width=1\linewidth]{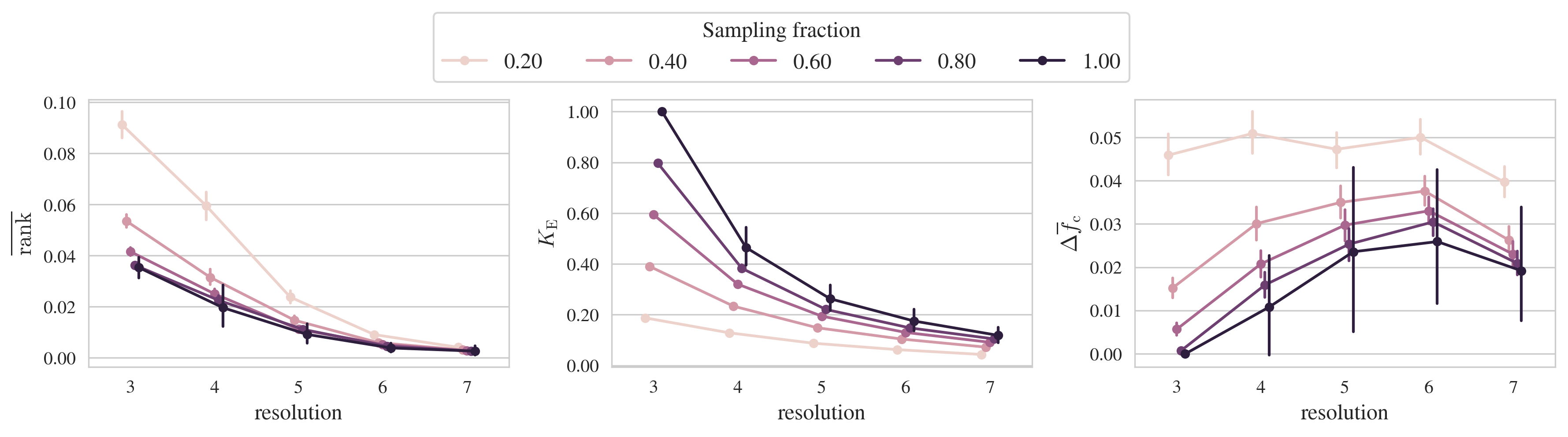}
    \caption{Performance of the heuristic search executing a partial exploration of the rule family partitions, segmented by sampling fraction, with 95\% confidence interval.}
    \label{fig:hs_performance_sampl}
\end{figure}

The performance of the heuristic search is compared with the exhaustive approach in two scenarios. First, we test its effectiveness in finding good classification rules when exploring whole rule family partitions (\autoref{fig:hs_performance_full}). Then, we evaluate the impact of the partial exploration in the heuristic search performance, by applying a sampling in the rule partitions (\autoref{fig:hs_performance_sampl}). For each sampling fraction, the heuristic search was repeated 25 times. Three metrics are used for this comparison:
\begin{itemize}
    \item \cmt{The rank of a rule $\varphi \in \Phi^\alpha$, given by $\overline{\text{rank}}(\varphi) = i/|\Phi^\alpha|$, denotes the chance of finding a rule of equal or higher accuracy, where $i \in \{1, \dots, 2^{2\alpha}\}$ means that $\varphi$ is the $i$-th best rule in that family;}
    \item The exploration ratio $K_\text{E}$ accounts for how many rules were evaluated since the beginning of the exploration divided by the size of the current rule family;
    \item The current accuracy error is the relative difference between the accuracy of the best current rule to the global optimum of the current rule family: $\Delta\bar{f}_c = 1 - f' / f^\alpha_\text{max}$
\end{itemize}

The first scenario shows the performance of the heuristic approach (see \autoref{fig:hs_performance_full}). As the resolution increases, the heuristic finds rules closer to the current global optimum (left plot) with a $K_\text{E}$ that decays exponentially and approaches 20\% on average (middle plot). Also, $\Delta\bar{f}_c$ indicates that even without finding the rule with best accuracy it can find competitive solutions for the rule optimization problem (right plot), sometimes even surpassing the result obtained by the exhaustive search ($\Delta\bar{f}_c < 0$). This unexpected behavior is explained by the accuracy versus resolution curve that is not strictly monotonic. As occurs for the github-stargazers and twitch-egos social datasets (\autoref{fig:acc_distrib_box_res}), the incremental optimization of ILS-based heuristics can find particular rules in smaller rule families that are better than all other rules in the immediately subsequent families.

The second scenario highlights the robustness of the method and the trade-off between computational complexity and classification performance (see \autoref{fig:hs_performance_sampl}). Even when we reduce the fraction of rules explored in the selected partitions and, consequently the exploration rate $K_\text{E}$, the resulting increase of $\Delta\bar{f}_c$ remains low. Notably, the method maintains $\Delta\bar{f}_c < 4\%$ despite the sampling fraction of 0.4, the case where $K_\text{E}$ approaches 10\% -- half of the value obtained in the first scenario.

\begin{figure}[!ht]
    \centering
    \includegraphics[width=0.8\linewidth]{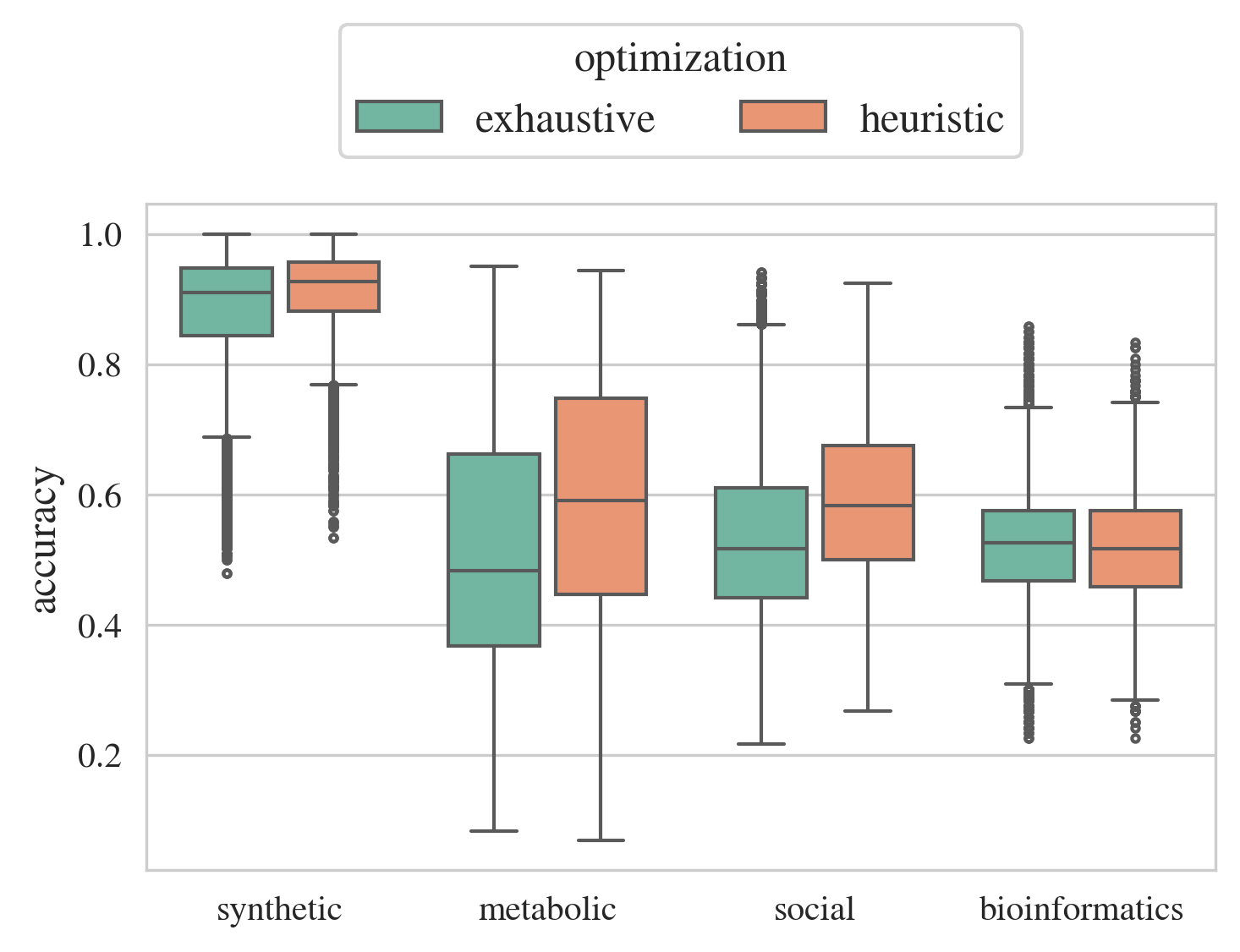}
    \caption{Accuracy distribution of rules explored by exhaustive search ($\alpha = 7$) versus heuristic search ($\alpha = 3, \dots, 7$) without sampling, grouped by dataset scope. The box spans the inter-quartile range, with the median shown as a horizontal line and outliers as individual points.}
    \label{fig:hs_performance_distrib}
\end{figure}

Ultimately, the rules explored by the heuristic strategy have better general performance than the exhaustive approach (see \autoref{fig:hs_performance_distrib}), resulting in a distribution with higher median accuracy for 22 of the 23 datasets ($p\text{-value} \le 0.05$).

\section{Conclusion}
\label{sec:conclusion}

The jaggedness is a quantitative parameter based on the visual notion of rule's shape irregularity, and provides valuable insight into the behavior of Life-Like Network Automata when classifying networks, guiding the understanding and prediction of rule accuracy.

From a theoretical standpoint, the jaggedness aligns with established measures such as Hamming weight and Boolean sensitivity, but distinguishes itself by offering an intuitive perspective on automaton dynamics that is independent of network topology. Although it may not seem to be the most descriptive metric, it effectively serves as a fundamental starting point to understand how much the automaton rule is a determining factor into the generation of different dynamics, being a better indicator of rule suitability for classification than the aforementioned measures.

In terms of computational efficiency, the reduction of the search space complexity brings feasibility to large-scale applications. By analyzing the accuracy of the rules according to their jaggedness at low resolution rule families and discovering where the best rules are, it is possible to extrapolate it to higher resolutions and gradually improve the rule quality, dispensing the need to scan the entire rule space. The experiments showed an efficiency gain of 10x as the exploration ratio approximates a fixed value ($K_\text{E} \to 10\%$), what could be improved even more by using more sophisticated local search procedures and hybrid metaheuristics.

Lastly, this work extends the findings about the dependence between the accuracy and the resolution of the rule space to more datasets, reinforcing the applicability of the LLNA as a method to characterize CNs in a wide range of contexts.

\section*{CRediT authorship contribution statement}
\label{sec:credit}

\textbf{Lucas Caldeira de Oliveira:} Conceptualization, Methodology, Software, Writing - Original Draft, Writing - Review \& Editing. \textbf{Michiel Rollier:} Methodology, Writing - Review \& Editing. \textbf{Jan M. Baetens:} Supervision, Writing - Review \& Editing. \textbf{Odemir M. Bruno:} Conceptualization, Funding acquisition, Supervision.

\section*{Declaration of competing interest}
\label{sec:declaration}

The authors declare that we have no known competing financial interests or personal relationships that could have appeared to influence the work reported in this paper.

\section*{Acknowledgments}
\label{sec:acknowl}

This work has been partially supported by the São Paulo Research Foundation (FAPESP) regular grant \#2024/02727-0 and internship abroad grant \#2024/16345-1; by the Brazilian Federal Agency for Support and Evaluation of Graduate Education (CAPES) grant \#88887.841805/2023-00; by the Research Foundation Flanders (FWO) grant with project title ``An analysis of network automata as models for biological and natural processes'' [3G0G0122]; by the FWO travel grant with file name V412625N; and by the FWO congress participation grant K105625N.


\bibliographystyle{ieeetr}
\bibliography{references}

\end{document}